\documentclass{sig-alternate-05-2015}

\usepackage{array,booktabs,makecell}
\usepackage{multirow}
\usepackage{amsmath}
\usepackage{bm}
\usepackage[linesnumbered,ruled]{algorithm2e}
\usepackage{mathtools}
\usepackage{dblfloatfix}
\usepackage{subfigure}
\usepackage[export]{adjustbox}
\usepackage{comment}
\usepackage{etoolbox}
\usepackage{enumerate}

\DeclarePairedDelimiter\ceil{\lceil}{\rceil}

\makeatletter
\def\@copyrightspace{\relax}
\makeatother

\usepackage[pagebackref=true,breaklinks=true,letterpaper=true,colorlinks,bookmarks=false]{hyperref}

\newbool{inccomment}
\setbool{inccomment}{true}
\newcommand{\hl}[1]{\ifbool{inccomment}{{\color{magenta}#1}}{}}
\newcommand{\cc}[1]{\ifbool{inccomment}{{\color{red}CC: #1}}{}}

\begin{document}
\title{Group Scissor: Scaling Neuromorphic Computing Design to Large Neural Networks}

\numberofauthors{5}
\author{
	%
	%
\alignauthor
Yandan Wang\\
	\affaddr{University of Pittsburgh}\\
	\email{yaw46@pitt.edu}	
\alignauthor
Wei Wen\\
\affaddr{University of Pittsburgh}\\
\email{wew57@pitt.edu}
\alignauthor
Beiye Liu\\
\affaddr{University of Pittsburgh}\\
\email{bel34@pitt.edu}
\and
\alignauthor
Donald Chiarulli\\
\affaddr{University of Pittsburgh}\\
\email{don@pitt.edu}
\alignauthor
Hai (Helen) Li\\
\affaddr{University of Pittsburgh}\\
\email{hal66@pitt.edu}
}

\maketitle
\begin{abstract}
Synapse crossbar is an elementary structure in Neuromorphic Computing Systems (NCS). However, the limited size of crossbars and heavy routing congestion impedes the NCS implementations of big neural networks. In this paper, we propose a two-step framework (namely, \textit{group scissor}) to scale NCS designs to big neural networks. The first step is \textit{rank clipping}, which integrates low-rank approximation into the training to reduce total crossbar area. The second step is \textit{group connection deletion}, which structurally prunes connections to reduce routing congestion between crossbars. Tested on convolutional neural networks of \textit{LeNet} on MNIST database and \textit{ConvNet} on CIFAR-10 database, our experiments show significant reduction of crossbar area and routing area in NCS designs. Without accuracy loss, rank clipping reduces total crossbar area to 13.62\% and 51.81\% in the NCS designs of \textit{LeNet} and \textit{ConvNet}, respectively. Following rank clipping,  group connection deletion further reduces the routing area of \textit{LeNet} and \textit{ConvNet} to  8.1\% and 52.06\%, respectively.
\end{abstract}

\section{Introduction}
The record-breaking classification performance of \textit{deep neural networks} (DNNs)~\cite{krizhevsky2012imagenet} in recent years has stimulated the fast-growing research on hardware design of \textit{neuromorphic computing systems} (NCS)~\cite{jo2010nanoscale}\cite{esser2013cognitive}\cite{xu2011design}\cite{hu2014memristor}\cite{li2014training}\cite{wen2016new}. 
NCS utilize devices and circuit components to mimic the behaviors of neural networks to perform intelligent tasks, such as image classification, speech recognition and natural language processing. 
Circuit-level and architecture-level NCS designs using emerging memristor devices~\cite{strukov2008missing} and traditional CMOS technologies~\cite{esser2013cognitive} are being explored.  

In software applications, the depth of DNNs rapidly grows from several layers to hundreds or even thousands of layers~\cite{he2015deep}.
However, the scale of hardware design of NCS falls far behind. 
A major critical issue that obstructs the scaling-up of NCS to big neural networks is the limited synaptic connection (e.g., crossbar) in hardware implementation.
Accordingly, it results in heavy wire congestion (e.g., the routing between crossbars). 
Taking the memristor-based NCS as an example, under the impact of IR-drop and process variations, both reading and writing reliability will be severely degraded when the size of a memristor-based crossbar is beyond $64\times64$~\cite{liang2010cross}\cite{liu2014reduction}. 
The similar scenario can also be observed in CMOS-based conventional designs. 
For example, the IBM TrueNorth chip, as a pioneer in NCS design, limits the size of neurosynaptic crossbars to $256 \times 256$~\cite{esser2013cognitive}. 
It is inevitable to interconnect multiple crossbars to implement modern big neural networks.
The increasing scale of neural networks could quickly exhaust the resources of synapse crossbars and deteriorate the wire congestion~\cite{wen2015eda}\cite{akopyan2015truenorth}.

To solve those problems, \cite{akopyan2015truenorth} mapped logically-connected cores to physically-adjacent cores to reduce spike communications. However, it only optimized the placement of cores and cannot reduce the core consumption.
The existing NCS optimization based on traditional sparse neural networks can alleviate the wire congestion~\cite{wen2015eda}. However, they usually separate the software sparsifying and hardware deployment, which makes the optimization more challenging. 

Unlike previous work, we propose a tow-step framework -- \textit{group scissor} -- to overcome above issues so as to scale NCS designs to big neural networks.
The first step is \textit{rank clipping}, which integrates low-rank approximation into the training process of neural networks. It targets at reducing the dimensions of connection arrays in a group-wise way and therefore reducing the consumption of synapse crossbars in NCS. 
The second step -- \textit{group connection deletion} -- structurally deletes/prunes groups of connections. 
The approach tends to learn hardware-friendly sparse neural networks to directly delete the routing wires between crossbars, with controllable low hardware cost. 
Unlike~\cite{wen2015eda} which evaluated NCS by Hopfield networks using less challenging database, we evaluate our \textit{group scissor} by state-of-the-art convolutional neural networks -- \textit{LeNet} and \textit{ConvNet} -- using MNIST and CIFAR-10 database. 
Our experiments show, without accuracy loss, \textit{rank clipping} respectively reduces total crossbar area to 13.62\% and 51.81\% in \textit{LeNet} and \textit{ConvNet}, and \textit{group connection deletion} reduces the routing area to  8.1\% and 52.06\%, respectively.

\begin{figure}[b] 
	\begin{center}
	\vspace{-12pt}
	\includegraphics[width=0.9\columnwidth]{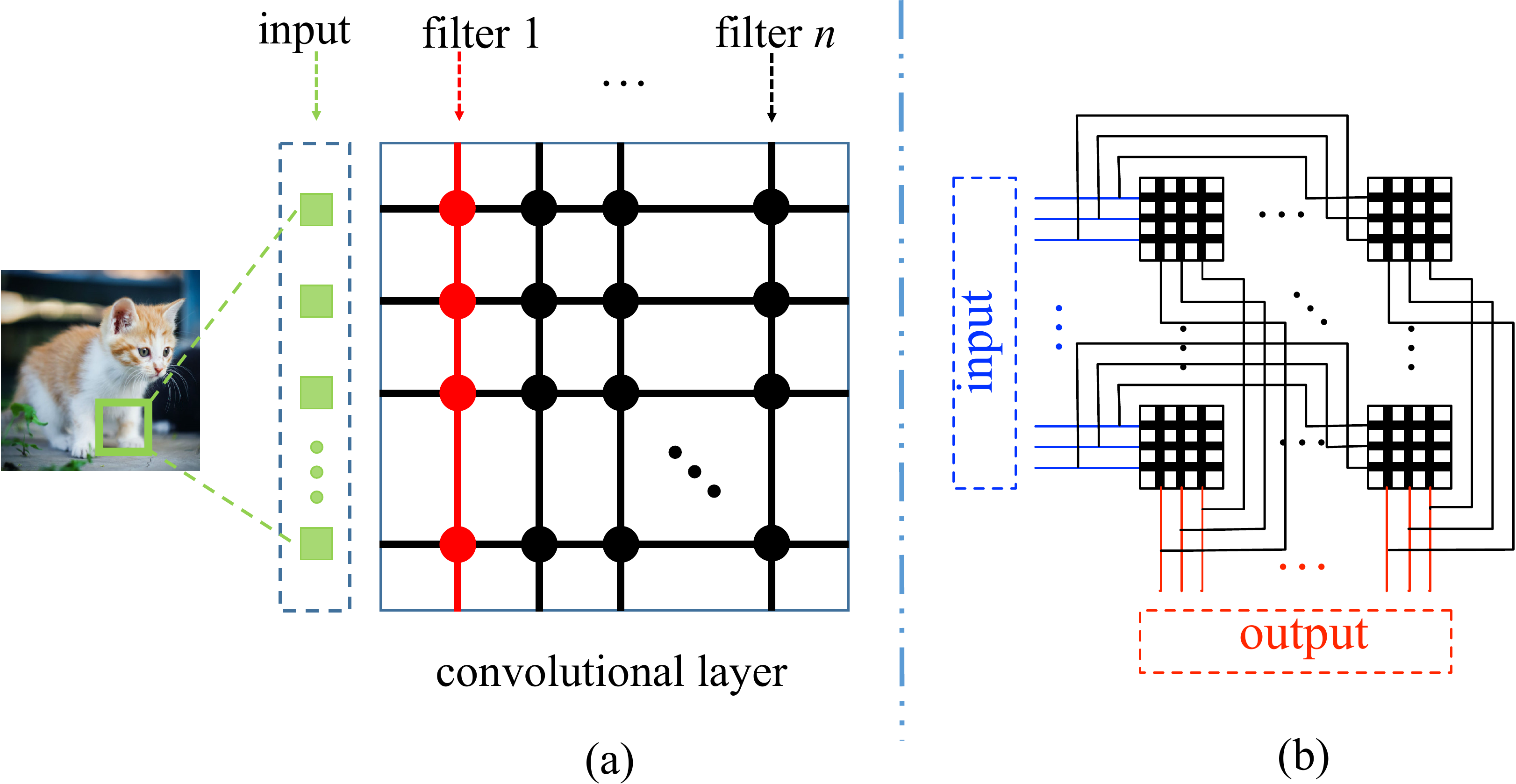}		
	\vspace{-9pt}
	\caption{The NCS designs for (a) a small convolutional layer, and (b) a big layer.} 
	\vspace{-12pt}
	\label{NN_xbar}
	\end{center}
\end{figure}

\vspace{-3pt}
\section{Preliminary}
\label{sec:preli}



Figure~\ref{NN_xbar}(a) illustrates an implementation of a convolutional layer in neural networks using \textit{memristor-based crossbars} (MBC), where memristors (a.k.a. synapses) in each column encode the weights of one filter \cite{linghao2017pipelayer}. 
The implementation of a fully-connected layer utilizes the similar structure, but each column realizes the connections to one output neuron. 
As aforementioned, the sizes of crossbars are limited.
So when implementing big neural networks, a high volume of the interconnection of crossbars are required.

Figure~\ref{NN_xbar}(b) depicts a circuit-level implementation of a big layer by tiling and interconnecting MBC~\cite{wen2015eda}. 
As the scale of modern neural networks grows, the high crossbar area occupation and heavy routing congestion become critical issues and seriously obstruct the scalability of the hardware implementation.

\section{The Group Scissor Framework}
\label{sec:metho} 

In this work, we propose the \textit{Group Scissor} framework to improve the scalability of neuromorphic computing design. 
The framework contains two steps: \textit{rank clipping} for crossbar area occupation reductions and \textit{group connection deletion} for routing congestion reduction. 
The details of the proposed design will be described in this section. 
Moreover, the estimations of circuit area and routing wires for MBC-based neuromorphic design are formulated.

\subsection{Rank Clipping}
\label{subsec:metho:pruning} 

As discussed above, the high crossbar area occupation and heavy routing congestion are the major obstacles in realizing big neural networks. 
In order to overcome these issues, we propose to utilize \textit{low-rank approximation} (LRA) to reduce the dimensions of weight (connection) matrices in big neural networks. 
Low-rank approximation is a mathematical technology, which uses the product of smaller matrices with reduced rank to approximate a given large matrix. Specifically, an original weight matrix $\textbf{W} \in~\mathbb{R}^{N \times M}$ can be approximated as
\begin{equation}
\textbf{W} \approx \textbf{U} \cdot \textbf{V}^T = \widetilde{\textbf{W}},
\end{equation}
where $\textbf{U} \in~\mathbb{R}^{N \times K}$, $\textbf{V}^T \in~\mathbb{R}^{K \times M}$, and $K$ is the rank of the approximation. When $K<<M$, $\textbf{U}$ and $\textbf{V}$ are reduced to skinny matrices. The total crossbar area occupation can be reduced when the rank $K$ satisfies 
\begin{equation}
K< \frac{NM}{N+M}.
\end{equation}

There are various LRA techniques. 
Without losing generality, commonly used \textit{principal components analysis} (PCA) \cite{jolliffe2002principal} and \textit{singular value decomposition} (SVD)~\cite{liu2014reduction} are adopted as the representatives in this work. 

\begin{algorithm}[t]
	\label{algo:pca}
	\small
	\SetKwInOut{Input}{Input}
	\SetKwInOut{Output}{Output}
	\caption{Principal Components Analysis (PCA)}
	\Input{$N \times M$ matrix $\textbf{W}$, and rank $K$}
	Get mean of rows $\textbf{w}_{n}~\forall n\in [1...N]$:  $\bm{\mu}=\frac{1}{N}\sum_{n=1}^{N}\textbf{w}_{n}$;\\
	Centralize the data: replace each $\textbf{w}_{n}$ with $\textbf{w}_{n}-\bm{\mu}$;\\
	Calculate the $M \times M$ covariance matrix: $\textbf{C}=\frac{\textbf{W}^{T}\textbf{W}}{N-1}$;\\
	Calculate the eigenvectors $\textbf{v}_{m}$ and eigenvalues $\lambda_m$ of covariance matrix $\textbf{C}$: $\textbf{C}\textbf{v}_{m}=\lambda_{m}\textbf{v}_{m}$, $\forall m \in [1...M]$;\\
	Project to subspace: $N\times K$ matrix $\textbf{U}=\textbf{W}\textbf{V}$, where $\textbf{V}=[{\textbf{v}_{1}},...{\textbf{v}_{K}}]$ is a $M\times K$ matrix and $\textbf{v}_{1...K}$ are eigenvectors corresponding to the largest $K$ eigenvalues;\\
	\Output{$N \times M$ approximation matrix $\widetilde{\textbf{W}} = $\textbf{U}$ \cdot \textbf{V}^T$}
\end{algorithm}

The PCA approach is formulated in Algorithm~\ref{algo:pca}.
The essence of PCA is a linear projection from a high dimensional space ($\textbf{w}_{n}\in\mathbb{R}^M$) to a lower dimensional subspace ($\textbf{u}_{n}\in\mathbb{R}^K$, $K\ll M$) to minimize the reconstruction error of $\textbf{W}$, where $\textbf{w}_{n}$ and $\textbf{u}_{n}$ is the $n$-th row of $\textbf{W}$ and $\textbf{U}$, respectively, and $\textbf{V}$ is the basis of the subspace. The reconstruction error is
\begin{equation}
e_K = \frac{|| \textbf{W} - \widetilde{\textbf{W}} ||^2 }{|| \textbf{W}||^2 }
= \frac{\sum_{m=K+1}^{M} \lambda_m}{\sum_{m=1}^{M} \lambda_m},
\end{equation}
where $||\cdot||$ is the Euclidean norm, namely Euclidean distance.


\begin{table*}[t]
	\caption{Accuracy and ranks} 
	\label{tab:accu_lra_vs_clipping}
	\centering
	\begin{tabular}{cccc|rccccc}
		\toprule [0.1 em]
		\toprule [0.1 em]
		\textbf{Database}& Net &\textbf{Method} & \textbf{Accuracy} & & conv1\textsuperscript{$\dagger$} & conv2 & conv3 & fc1\textsuperscript{$\dagger$} & fc2 \\ 
		\midrule [0.07 em]
		\multirow{3}{*}{MNIST} & \multirow{3}{*}{\textit{LeNet}~\cite{lecun1998gradient}} & \textit{Original}\textsuperscript{$\ddagger$} & 99.15\% & \multirow{3}{*}{Rank $K$} & 20 & 50 & -- & 500 & 10\\ 
		&& \textit{Direct LRA} & 96.44\% & & 5 & 12 & -- & 36 & 10\\ 
		&& \textit{Rank clipping} & 99.14\% & & 5 & 12 & -- & 36 & 10\\  
		
		\midrule [0.07 em]
		\multirow{3}{*}{CIFAR-10} & \multirow{3}{*}{\textit{ConvNet}~\cite{krizhevsky2012imagenet}} & \textit{Original} & 82.01\% & \multirow{3}{*}{Rank $K$} & 32 & 32 & 64 & 10 & --\\
		&& \textit{Direct LRA} & 43.29\% & & 12 & 19 & 22 & 10 & --\\
		&& \textit{Rank clipping} & 82.09\% & & 12 & 19 & 22 & 10 & --\\
		
		\bottomrule [0.1 em]
		\bottomrule [0.1 em]
		\multicolumn{10}{l}{\textsuperscript{$\dagger$} conv1 is the first convolutional layer, fc1 is the first fully-connected layer, and so forth} \\
		\multicolumn{10}{l}{\textsuperscript{$\ddagger$} corresponding rank indicates the number of filters in convolutional layers or indicates the number of output}\\
		\multicolumn{10}{l}{~  neurons in fully-connected layers.}
	\end{tabular}
\end{table*}

\begin{figure}[b]
	\centering
	\vspace{-9pt}
	\includegraphics[width=0.95\columnwidth]{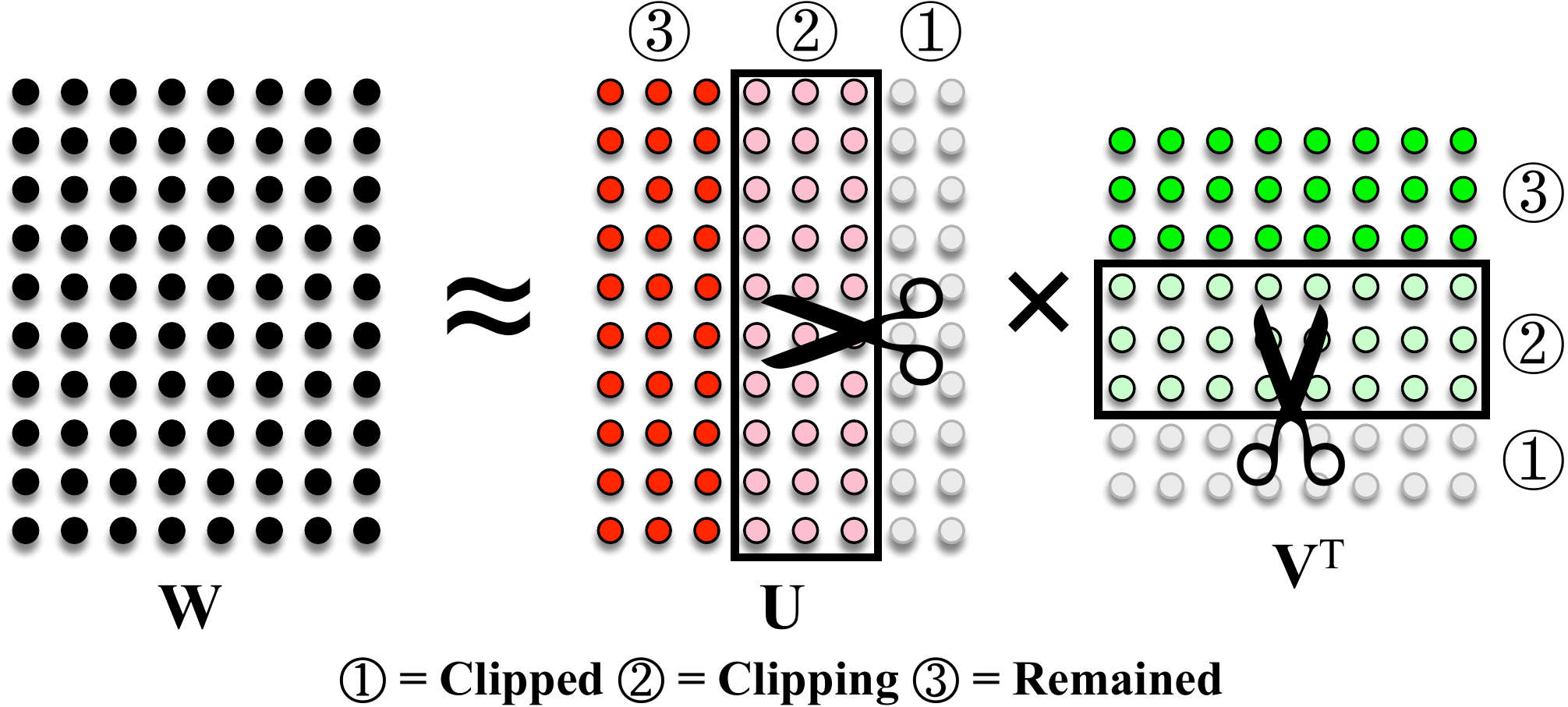}	
	\vspace{-9pt}
	\caption{Rank clipping for crossbar area occupation reduction.}	
	\label{fig:rank_clipping}
\end{figure}

Although LRA can approximately reconstruct the original weights, small perturbation of weights can deteriorate the classification accuracy. 
Table~\ref{tab:accu_lra_vs_clipping} compares the performance of the original baseline design (``\textit{Original}'') and the low-rank neural networks which are directly decomposed by PCA (``\textit{Direct LRA}''). 
The accuracy drops rapidly after applying ``\textit{Direct LRA}''. 
Fine-tuning (retraining) the low-rank neural networks can recover accuracy, but the optimal ranks in all layers are unknown.
More importantly, it is very time-consuming to explore the entire design space by decomposing and retraining a wide variety of neural networks. 
We propose the LRA-based \textit{rank clipping}, which can not only successfully retain the accuracy but also can automatically converge to the optimal low ranks in all layers. 
Low ranks in Table~\ref{tab:accu_lra_vs_clipping} are actually obtained by our rank clipping method.

The key idea of rank clipping is illustrated in Figure~\ref{fig:rank_clipping}. 
Rather than direct LRA after training, we integrate LRA into the training process and carefully clip some ranks with small reconstruction errors after a fixed number of training iterations, say,  $S$ iterations. 
The gentle clipping induces small reconstruction errors and thus slightly affect the classification accuracy.
As such, the accuracy could be recovered by the following $S$ iterations. 
The iteration of clipping and training not only avoids irremediable accuracy degradation but also enables neural networks to gradually converge to the optimal ranks for all layers.


\begin{algorithm}[t]
	\label{algo:MBC_size_clipping_method}	
	\small
	\SetKwInOut{Input}{Input}
	\SetKwInOut{Output}{Output}
	\caption{Rank Clipping} 
	\Input{Trained original neural network, tolerable clipping error $\varepsilon$, maximum training iteration $I$, clipping step $S$}
	\For{each layer l}
	{
		PCA of weight matrix $\textbf{W}_l = \textbf{U}_l \cdot \textbf{V}^T_l$ with full rank $K_l=M_l$;  \\
		}
	\While{$i=1; i < I; i=i+S$}
	{
		\For{each layer l}
		{
			PCA of $\textbf{U}_l = \hat{\textbf{U}}_l \cdot \hat{\textbf{V}}^T_l $ using the minimum rank $\hat{K}$ which satisfies $e_{\hat{K}} \leq \varepsilon$; \\
			
			\eIf{$\hat{K} < K_l$}{
				$K_l = \hat{K}$;  $\textbf{U}_l = \hat{\textbf{U}}_l$;  $\textbf{V}^T_l = \hat{\textbf{V}}^T_l \cdot  \textbf{V}^T_l $
			}{
			continue;
			}
		}
		
		Train the neural network for $S$ iterations;
	}
	\Output{Clipped low-rank neural network with approximation $\textbf{W}_l = \textbf{U}_l \cdot \textbf{V}^T_l$ for each layer $l$ }
\end{algorithm}


\begin{figure}[b]
	\centering
	\includegraphics[width=0.8\columnwidth]{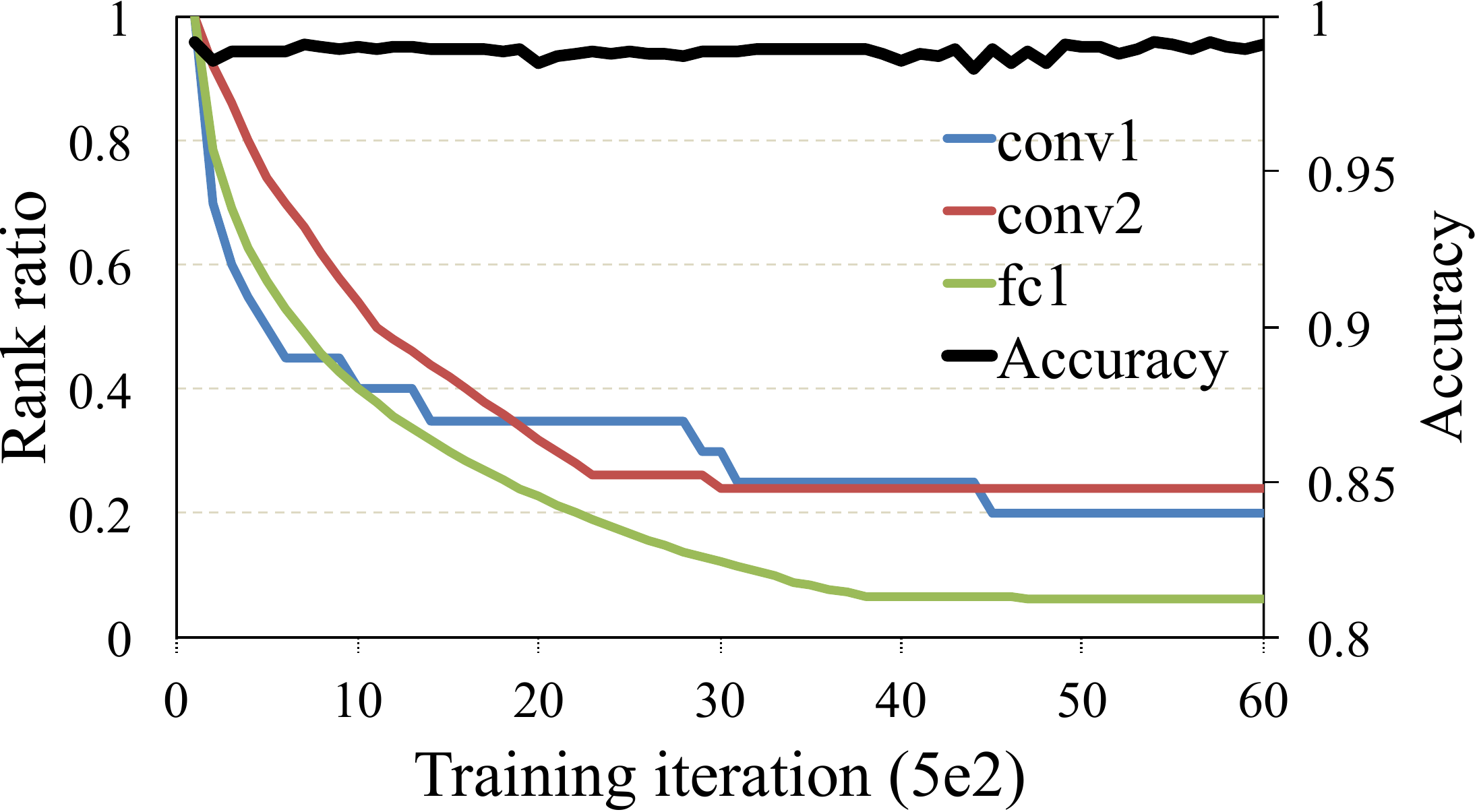}	
	\vspace{-12pt}
	\caption{Rank ratio of each layer and accuracy during training with rank clipping.}	
	\label{fig:rank_clipping_accuracy_rank_ratio}
	\vspace{-12pt}
\end{figure}

Algorithm~\ref{algo:MBC_size_clipping_method} describes the detailed operation of the rank clipping.
The \textit{tolerable clipping error} $\varepsilon$ is the allowed maximum reconstruction error after each rank clipping. 
A gentle clipping can be enabled by setting a small $\varepsilon$, \textit{e.g.}, $0.01$. 
Our rank clipping starts with a full-rank LRA without reconstruction errors, and iteratively examines if the low-dimensional $\textbf{U}$ can be further projected to a \textit{lower}-rank subspace with only reconstruction error of $\varepsilon$.
Note that PCA is used as the representative of LRA in Algorithm~\ref{algo:MBC_size_clipping_method}.
However, other LRA methods like SVD can also be used. 
The only modification is to replace the approximation of weight matrix by the other LRA method.

Figure~\ref{fig:rank_clipping_accuracy_rank_ratio} plots the trends of rank reduction and accuracy retention during the rank clipping of \textit{LeNet} in Table~\ref{tab:accu_lra_vs_clipping} using PCA. 
Rank clipping is examined every $S=500$ (denoted as 5e2 in x-axis title) iterations with $\varepsilon=0.03$. 
In the figure, the \textit{rank ratio} is defined as the remained rank over full rank, i.e., $K/M$. 
The figure demonstrates that ranks are rapidly clipped at the beginning of iterations and converge to optimal low ranks.
During the entire process, the accuracy changes are limited to small fluctuations.
 
As shown in Figure~\ref{fig:rank_clipping_accuracy_rank_ratio} and Table~\ref{tab:accu_lra_vs_clipping}, the proposed rank clipping successfully reduces the ranks in both convolutional layers and fully-connected layers without accuracy loss. 
The crossbar area occupation of the entire \textit{LeNet} (\textit{ConvNet}) reduces to 13.62\% (51.81\%).
Instead of PCA, when SVD is applied, the whole crossbar area can also be reduced to 32.97\% (55.64\%) for \textit{LeNet} (\textit{ConvNet}), which indicates SVD is inferior to PCA. Therefore, we mainly conduct experiments using PCA approach.
Note that the last layers of \textit{LeNet} and \textit{ConvNet} are not clipped because the rank ($M=10$) is already very small and there is little improvement space. 

\subsection{Group Connection Deletion}
\label{subsec:metho:rout} 
 
\begin{figure}[t]
   	\begin{center}
   		\includegraphics[width=0.61\columnwidth]{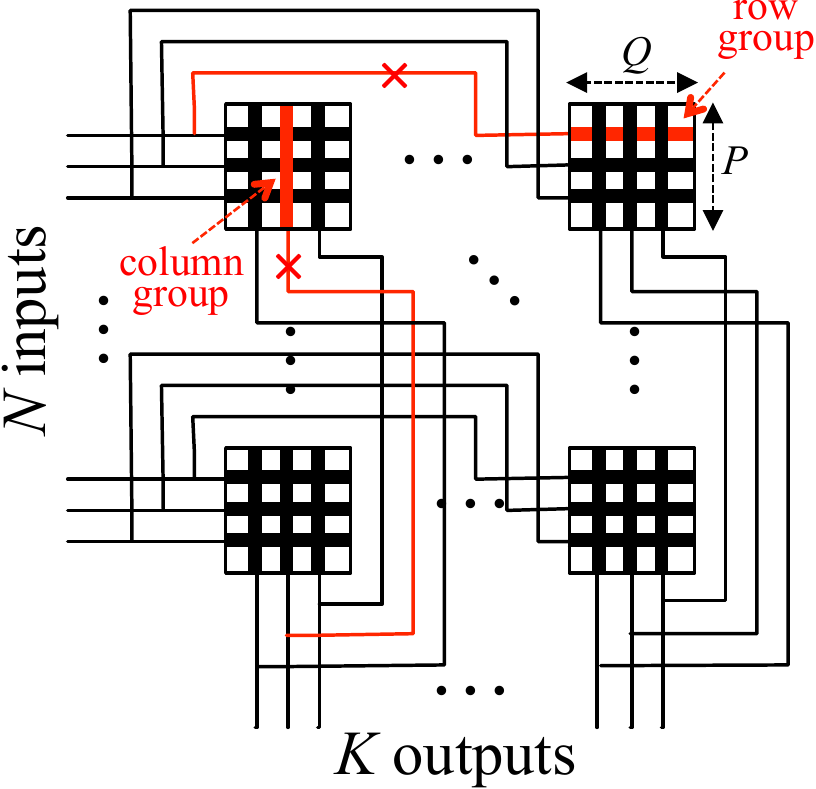}		
   		\vspace{-9pt}
   		\caption{The group connection deletion.} 
   		\label{routing_clipping}
   		\vspace{-18pt}
   	\end{center}
\end{figure}
   
The rank clipping can reduce the total number of required crossbars, but a large number of crossbars will be still necessary to implement modern big neural networks. 
The second step of our \textit{group scissor} framework -- \textit{group connection deletion} aims at removing interconnections between synapse crossbars so as to reduce the circuit-level routing congestion and architecture-level inter-core communication for NCS. 
  
Figure~\ref{routing_clipping} gives the basic idea of \textit{group connection deletion}. An array of MBC are interconnected to implement a large weight matrix $\textbf{U} \in~\mathbb{R}^{N \times K}$. 
Suppose the elementary synapse crossbar has $P$ inputs and $Q$ outputs $(P \ll N, Q \ll K)$, a $ \ceil*{\frac{N}{P}} \times \ceil*{\frac{K}{Q}} $ array of crossbars must be interconnected to implement \textbf{U} as illustrated in Figure~\ref{routing_clipping}. 
The implementation of another matrix \textbf{V} shall follow the similar method.
As memristors can be densely manufactured in the crossbar and the area of each memristor cell is feature-size level, the routing wires dominate the circuit area~\cite{wen2015eda}. 
Suppose a \textit{row group} of connections in Figure~\ref{routing_clipping} all have zero weights, implying that those connections are removable, we can delete/prune the wire routing to the input of this \textit{row group}. 
Similarly, the wire routing from the output of a \textit{column group} can be deleted when the \textit{column group} of connections are all-zeros.
Our \textit{group connection deletion} method actively deletes those groups of connections during the learning of neural networks, meanwhile maintaining the classification accuracy at the similar level.
 
We harness \textit{group Lasso regularization} to delete groups of connections. Group Lasso is an efficient regularization in the study of structured sparsity learning~\cite{GroupLasso_2006}\cite{wen2016learning}. With group Lasso regularization on each group of weights, a high percentage of groups can be regularized to all-zeros.
In our \textit{group connection deletion} method, weights are split to row groups and column groups as illustrated in the figure. 
And group Lasso regularization is enforced on each group. 
Mathematically, the minimization function for training neural network with group Lasso can be formulated as:
\begin{equation}
\label{eq:grouplasso_gcd}
E(\mathcal{W})=E_{D}(\mathcal{W})+\lambda \cdot \left( \sum_{g=1}^{G^{(r)}} || {\mathcal{W}^{(r)}_g} || + \sum_{g=1}^{G^{(c)}} || {\mathcal{W}^{(c)}_g} || \right),
\end{equation} 
where $\mathcal{W}$ is the set of weights in the whole neural network, 
$E_{D}(\mathcal{W})$ is the original minimization function when training traditional neural networks.
$G^{(r)}$ and $G^{(c)}$ respectively denote the number of row groups and column groups, 
and $\mathcal{W}^{(r)}_g$ and $\mathcal{W}^{(c)}_g$ are the sets of weights in the $g$-th row group and column group, respectively. 
And
\begin{equation}
 \bigcup_{g=1}^{G^{(r)}} \mathcal{W}^{(r)}_g 
 = \bigcup_{g=1}^{G^{(c)}} \mathcal{W}^{(c)}_g 
 = \mathcal{W}.
\end{equation}
$\lambda$ is the hyper-parameter to control the trade-off between classification accuracy and routing congestion reduction. 
A larger $\lambda$ can result in lower accuracy but larger reduction of routing wires. 
During the back-propagation training with Eq.~(\ref{eq:grouplasso_gcd}), each weight $w$ will be updated as
\begin{equation}
\label{eq:weight_updating}
 w \leftarrow w - \eta \left( \frac{\partial E_{D}(\mathcal{W}) }{\partial w} + 
  \frac{ \lambda w}{||{\mathcal{W}^{(r)}_{i}}||}  + 
  \frac{ \lambda w}{||{\mathcal{W}^{(c)}_{j}}||} \right),
\end{equation} 
where $\eta$ is the learning rate, $i \in [1...G^{(r)}]$, $j \in [1...G^{(c)}]$, $w \in \mathcal{W}^{(r)}_{i}$ and $ w \in \mathcal{W}^{(c)}_{j} $.

With group connection deletion, we disconnect all the zero-weighted connections and prune all the routing wires connecting to all-zero row groups or column groups. 
After deletion, we fine-tune (retrain) the structurally-sparse neural networks to improve accuracy.

\begin{figure}
	\begin{center}
		\includegraphics[width=.8\columnwidth]{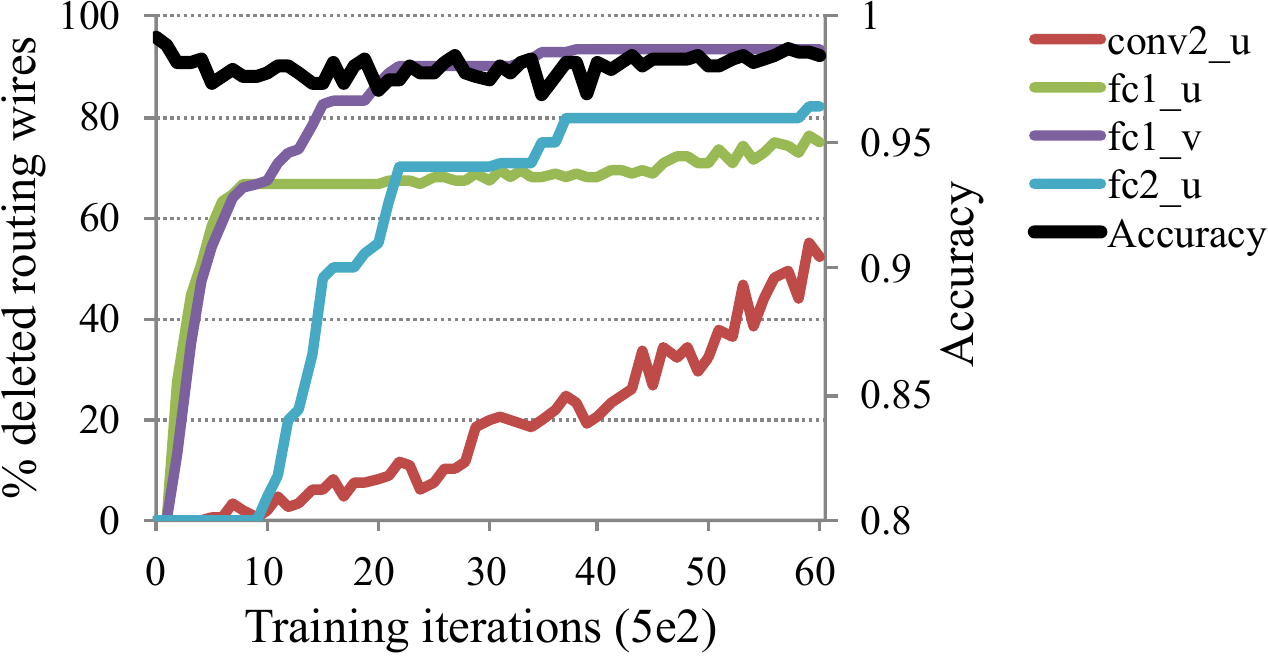}	
		\vspace{-10pt}	
		\caption{The percentage of deleted routing wires and accuracy during group connection deletion. fc1\_u and fc1\_v is the low-rank matrix \textbf{U} and \textbf{V} of fc1 after rank clipping, and so forth.} 
		\vspace{-15pt}
		\label{fig:iter_gcd}
	\end{center}
\end{figure}

Figure~\ref{fig:iter_gcd} plots the trends of deleted routing wires (i.e., all-zero row/column groups) and the classification accuracy versus the iterations of group connection deletion. 
The deletion process starts with the low-rank \textit{LeNet} in Table~\ref{tab:accu_lra_vs_clipping} that was already compressed by rank clipping. 
In Figure~\ref{fig:iter_gcd}, we only delete the matrices of \textbf{U} and \textbf{V} whose dimensions are beyond the largest size of MBC. 
More design details shall be presented in Section~\ref{subsec:metho:area} and Section~\ref{sec:expe}. 
Even for low-rank neural networks, our method can delete the routing wires dramatically,
\textit{e.g.}, 93.9\% interconnection wires are removed in the crossbar array of fc1\_v. Fine-tuning the deleted neural networks attains the baseline accuracy (99.1\%), 

Note that compared with our method, it is more challenging to use traditional sparse neural networks to reduce the routing wires. 
This is because its sparse weights are randomly distributed in the crossbar arrays and the corresponding routing wire must be preserved as long as there is one nonzero weight existing in the row group or column group.

\subsection{Area Estimation}
\label{subsec:metho:area} 

In this section, we formulate the area estimation method adopted for hardware evaluation in this work.

\vspace{2pt}
\textbf{\textit{MBC area estimation}}: 
The use of MBC in NCS design has been extensively studied. 
As a critical component in such a system, MBCs occupy a significant proportion of whole design area. 
Each MBC is an ultra high density cross-point structure formed by a set of memristors and wires. 
The area of a memristor cell in MBC is 4$F^{2}$ under the state-of-the-art technology~\cite{liu2014reduction}, where $F$ is the minimum feature size. 
Restricted by the technology limitations, a feasible MBC implementation only considers MBCs that are not larger than 64$\times$64~\cite{liang2010cross}.
To ensure the system reliability and robustness, 
we only consider MBCs with dimensions constrained within 64$\times$64 in the standard library.
For those large weight matrices in neural networks, their connections can be distributed into many MBCs in the library as demonstrated in Figure~\ref{NN_xbar}.

\textbf{\textit{Routing area estimation}}: 
Assume that the metal width is $W_{m}$, the distance between two metals is $W_{d}$, and the length of $i$-th wire between crossbars is $L_i$. 
The total routing area occupied by the wires can be roughly formulated as
\begin{equation}
\label{equation:metho:routing_area} 
A_{r}=(W_{m}+W_{d}) \sum_{i}^{N_{w}}L_i.
\end{equation}
Here $N_{w}$ is total wire count including electrostatic shielding wires. 
Suppose the average wire length is linearly proportional to $N_{w}$,
the routing area is estimated as
 \begin{equation}
 \label{equation:metho:routing_area_estimation} 
 A_{r}=\alpha N_{w}^2,
 \end{equation}
where $\alpha$ is a scalar.


\section{Experiment}
\label{sec:expe} 

In this section, experiments are conducted to evaluate the effectiveness of proposed rank clipping and group connection deletion methods. 
All the experiments conducted in this paper are based on the NCS implemented by MBC. 
The related experiment parameters on memristor and MBC are summarized in Table~\ref{tab:exper_para}. 
We mainly implement two neural networks -- $LeNet$ on MNIST and $ConvNet$ on CIFAR-10.
The detailed network structures are summarized in Table~\ref{tab:accu_lra_vs_clipping}.

\begin{table}
	\caption{ Experiment Parameters} 
	\label{tab:exper_para}
	\centering
	\begin{tabular}{c c}
		\toprule [0.1 em]
		\toprule [0.1 em]
		\parbox[c]{1mm}\textbf{Parameter} & \textbf{value}\\
		\midrule [0.07 em]
		memristor cell area&4$F^{2}$\\
		\midrule [0.07 em]
		maximum crossbar size&64$\times$64\\
		\midrule [0.07 em]
		Wire length between two memristors&2$F$\\
		\bottomrule [0.1 em]
		\bottomrule [0.1 em]
	\end{tabular}
\end{table}

\subsection{MBC Area Reduction}
\label{subsec:expri:MBC} 

\begin{figure}[b]
	\begin{center}
		\includegraphics[width=0.65\columnwidth]{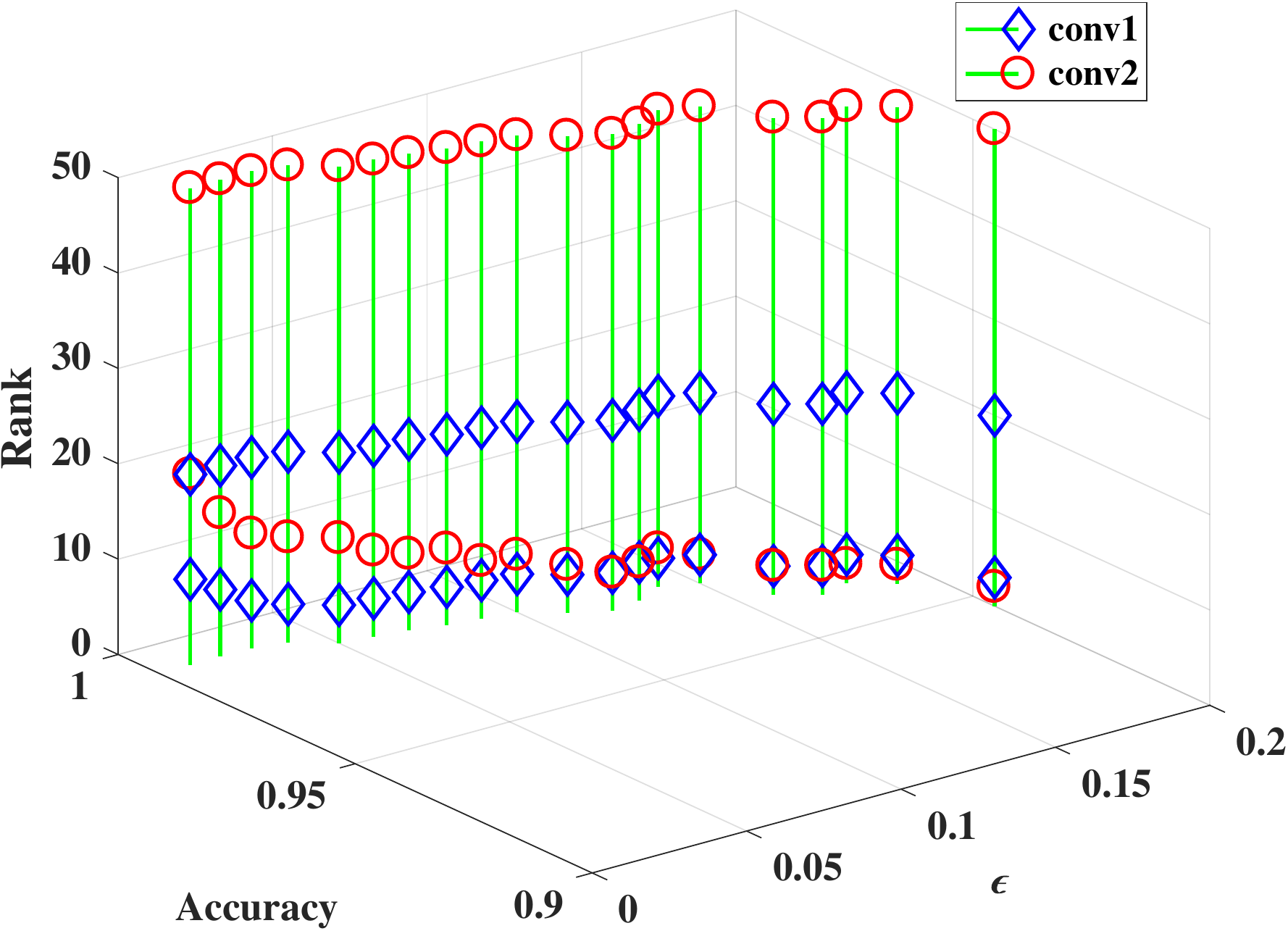}		
		\vspace{-9pt}
		\caption{The remained ranks in convolutional layers of \textit{Lenet}. fc1 is omitted for better visualization as its original rank $500$ is out of chart.} 
		\label{fig:pca_rank_reduction}
		\vspace{-15pt}
	\end{center}
\end{figure}

In our experiments, we clip all the convolutional and fully-connected layers, except the last classifier layer.
The original rank in the last layer is determined by the number of classes so the further reduction is meaningless.
The rank clipping method compresses each large weight matrix to two skinny matrices by reducing the rank.
Figure~\ref{fig:pca_rank_reduction} shows the final remained ranks with respect to the \textit{accuracy} and \textit{tolerable clipping error}~$\varepsilon$ for convolutional layers in $\textit{LeNet}$. 
Here the original rank of conv1 and conv2 is 20 and 50, respectively, as denoted by upper markers on the stems. 
For each layer, the rank decreases as $\epsilon$ increases, and finally reaches to a very small value.
It can be seen that the corresponding accuracy is well maintained. 
We also observe similar results in fc1. 
More specifically, the layer-wise ranks are reduced to $5$, $12$ and $36$ without accuracy loss, and to $4$, $6$ and $6$ with merely 1\% loss.

\begin{figure}[t]
	\begin{center}
		\includegraphics[width=1\columnwidth]{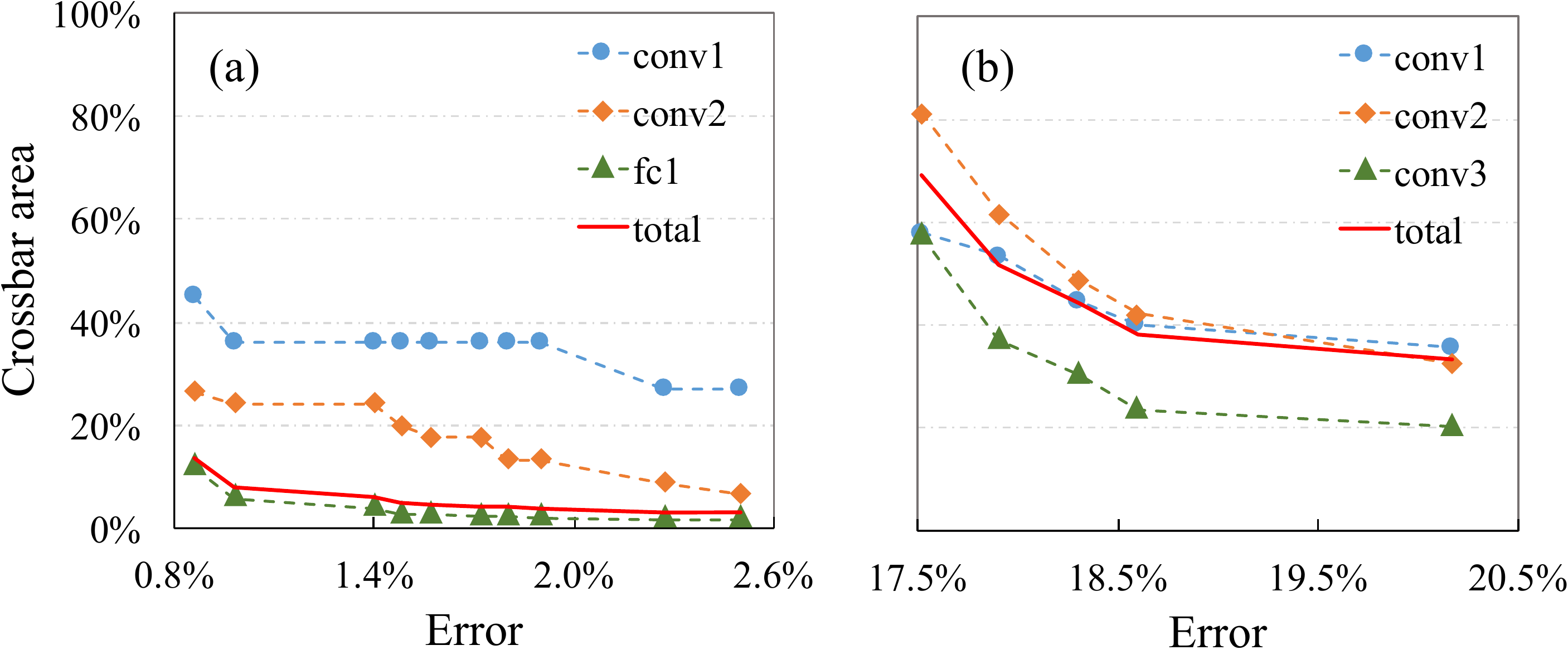}		
		\caption{The MBC area for (a) \textit{Lenet} and (b) \textit{ConvNet}, after applying the rank clipping.} 
		\label{fig:lenet_convnet_area}
		\vspace{-15pt}
	\end{center}
\end{figure}

Figure~\ref{fig:lenet_convnet_area}(a) and (b) respectively plot the percentage of remained MBC area with respect to the classification error for $LeNet$ and $ConvNet$. Routing area is excluded in this evaluation.
The area of each layer is the sum of the areas of \textbf{U} and \textbf{V}.
Total area includes the area of the last classifier layer, \textit{i.e.,} fc2 in \textit{Lenet} or fc1 in \textit{ConvNet}. 
For both networks, the layer-wise areas of both convolutional layers and fully-connected layers rapidly reduce with small accuracy loss. 

In summary, the rank clipping can reduce the total crossbar area of \textit{LeNet} to 13.62\% without sacrificing any accuracy loss. 
The crossbar area can be further reduced to 3.78\% with merely 1\% accuracy loss. 
For more challenging $ConvNet$, no accuracy loss is observed when the crossbar area is reduced to 51.81\%. 
And under an accuracy loss of 1\%, the total crossbar area can be reduced to 38.14\%.


\subsection{Routing Area Reduction}
\label{subsec:expri:rout} 

\begin{table}[t]
	\caption{MBC Sizes and remained routing wires in big layers.} 
	\label{tab:group_size}
	\centering
	\resizebox{.48\textwidth}{!}{
	\begin{tabular}{cccccccc}
		\toprule [0.1 em]
		\toprule [0.1 em]
		Net & type & conv1\_u & conv2\_{u} & conv3\_{u} & fc1\_{u} & fc1\_{v} & fc\_last \\ 
		\midrule [0.07 em]
		\multirow{2}{*}{\textit{LeNet}} & sizes & -- \textsuperscript{$\dagger$} & $50 \times 12$ & -- & $50 \times 36$ & $36 \times 50$ & $50 \times 10$\\ 
		&\% wires & -- & 47.5  & -- & 24.8 & 6.7 & 18.0\\
		
		\midrule [0.07 em]
		\multirow{2}{*}{\textit{ConvNet}} & sizes & $25 \times 12$ & $50 \times 19$ & $50 \times 22$ & -- & --& $64 \times 10$ \\
		&\% wires& 83.3 & 40.5  & 74.4 & -- & -- & 81.9\\
		\bottomrule [0.1 em]
		\bottomrule [0.1 em]
		\multicolumn{8}{l}{\textsuperscript{$\dagger$} The weight matrix can be implemented by one crossbar. }\\
		\multicolumn{8}{l}{ ~~~conv1\_v, conv2\_v and conv3\_v are omitted for the same reason. }\\
	\end{tabular}
	}
\end{table}

\begin{figure}[b]
	\begin{center}
		\vspace{-9pt}
		\includegraphics[width=1\columnwidth]{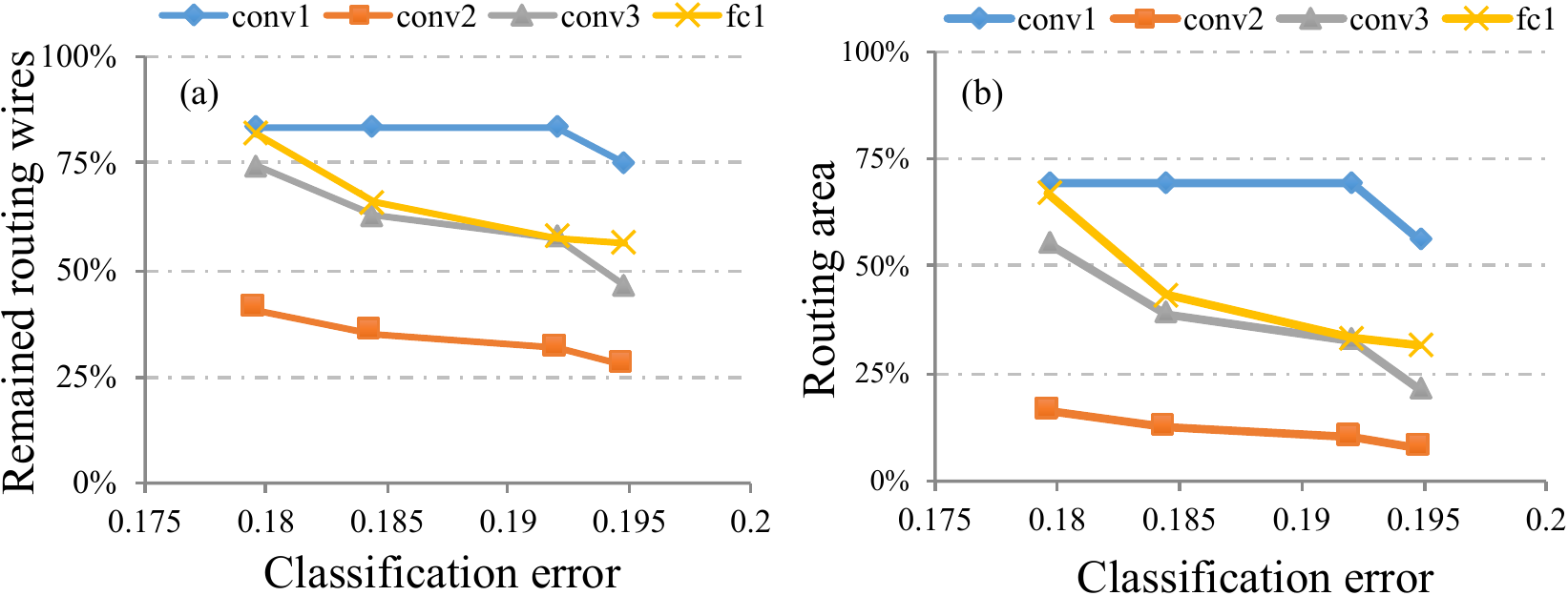}		
		\vspace{-18pt}
		\caption{The (a) routing wire and (b) routing area w.r.t. the classification error in \textit{ConvNet}. } 
		\label{fig:cifar_routing}
	\end{center}
\end{figure}

\begin{figure*}
	\vspace{-6pt}
	\subfigure
	{
		\includegraphics[width=0.1465\columnwidth,left]{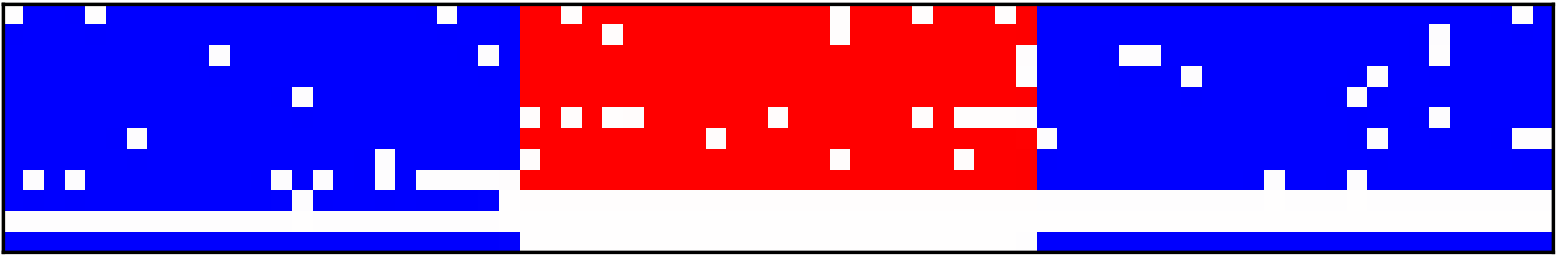}
	}
	\vfill
	\vspace{-5pt}
	\subfigure
	{
		\includegraphics[width=1.5625\columnwidth,left]{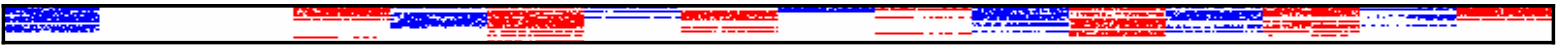}
	}
	\vfill
	\vspace{-5pt}
	\subfigure
	{
		\includegraphics[width=1.5625\columnwidth,left]{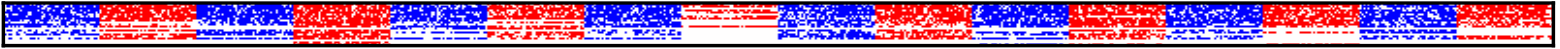}
	}
	\vfill
	\vspace{-5pt}
	\subfigure
	{
		\includegraphics[width=2.0\columnwidth,left]{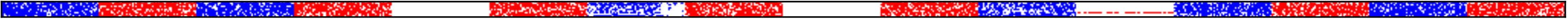}
	}
	\vspace{-15pt}
	\caption{ Weight matrices (transposed) after group connection deletion. The deletion starts from the rank-clipped \textit{ConvNet} in Table~\ref{tab:accu_lra_vs_clipping}. Matrices are plotted in scale in the order of conv1\_u, conv2\_u, conv3\_u and fc1. White regions have no connections. And connections in each blue/red block are implemented in a crossbar.} 
	
	\label{fig:lenet_conv1_filters}
\end{figure*}

To evaluate the routing congestion alleviated by group connection deletion method, we use the number of routing wires and remained routing area of Eq.~(\ref{equation:metho:routing_area_estimation}) as our metrics. 
Although the estimation of routing area in the real circuit can be more complex, the real routing area reduction in the hardware must be positively correlated to our results. 

As aforementioned in Section~\ref{subsec:metho:area}, our standard library contains all types of memristor crossbars with dimensions constrained within $64\times64$.
When implementing a $ N \times K $ weight matrix \textbf{U}, the MBC sizes are selected based on the following criteria:
(1)~Implement \textbf{U} in a $ N \times K $ MBC, when $N \le 64$ and $K \le 64$;
(2)~Implement \textbf{U} by an array of MBCs when $N > 64$ or $K > 64$, with the largest available MBC size $ P \times Q $, where $N$ and $K$ is divisible by $P$ and $Q$, respectively.

In the experiments, the group connection deletion starts with the rank-clipped \textit{LeNet} or \textit{ConvNet} without accuracy loss as presented in Table~\ref{tab:accu_lra_vs_clipping}. 
Based on the MBC selection criteria, the sizes of MBC utilized in big layers are shown in Table~\ref{tab:group_size}. 
Matrices with sizes constrained by $64 \times 64$ are omitted in the table, and no group Lasso regularization is enforced on those small matrices.

The experimental results of the remained routing wires after applying the group connection deletion without allowing accuracy loss are also presented in Table~\ref{tab:group_size}. 
The results for \textit{LeNet} are remarkable. We can achieve the same accuracy of the baseline, with routing wires being only 47.5\%, 24.8\%, 6.7\% and 18.0\% of the original ones in respective layer. 
This can reduce the layer-wise routing area to 8.1\%, on average.

Table~\ref{tab:group_size} also shows that, in \textit{ConvNet}, our method on average reduces layer-wise routing wires to 70.03\% and thus reduce layer-wise routing areas to 52.06\%, meanwhile achieving the same accuracy as the baseline. 
With an acceptable accuracy loss, the routing congestion can also be significantly alleviated.
Figure~\ref{fig:cifar_routing} comprehensively studies the remained routing wires and routing area under different classification errors. 
With merely 1.5\% accuracy loss, the routing area in each layer is reduced to 56.25\%, 7.64\%,  21.44\% and 31.64\%, respectively.

At last, Figure~\ref{fig:lenet_conv1_filters} shows the sparse weight matrices after group connection deletion for \textit{ConvNet} in Table~\ref{tab:group_size} without accuracy loss.
Each blue/red block stands for a collection of weights, which are implemented by one crossbar in the NCS design. 
White regions indicate that there are no connections.
After applying the group connection deletion, the connections in crossbars become sparse.
More importantly, the sparsity is structural instead of being randomly distributed in traditional sparse neural networks.
In the figure, a high ratio of column groups in crossbars are regularized to all-zeros, such that interconnection wires routing from those crossbar columns can be removed.
Impressively, as conv2\_u and fc1 in the figure show, some blocks have no connections in the whole region, indicating that the entire crossbar can be removed in the NCS implementation. 
It is significant because not only routing congestion can be alleviated, but also crossbar area can be reduced.    
We also note that a crossbar with some zero columns/rows can be replaced by a smaller but dense crossbar after removing those zero groups, which can further reduce the crossbar area.

\section{Conclusions}
\label{sec:conclusion} 

In this work, we propose a framework named \textit{group scissor} to alleviate the impact of hardware limitations on the NCS implementation of big neural networks. 
Specifically, \textit{rank clipping} and \textit{group connection deletion} methods are proposed to reduce area consumption of synapse crossbars and routing area between crossbars, respectively. Final experiments show that our methods can reduce crossbar area (routing area) to 13.62\% (8.1\%) with no accuracy loss for \textit{LeNet}. Moreover, no accuracy loss is observed for  more challenging \textit{ConvNet} when crossbar area is reduced to 51.81\% and routing area is reduced to 52.06\%. The proposed framework can significantly save hardware area and improve system scalability.

\bibliographystyle{ieeetr}
\small
\bibliography{sigproc}  

\end{document}